\documentclass{article}

    \usepackage[preprint]{neurips_data_2024}

\usepackage[utf8]{inputenc} %
\usepackage[T1]{fontenc}    %
\usepackage{hyperref}       %
\usepackage{url}            %
\usepackage{booktabs}       %
\usepackage{amsfonts}       %
\usepackage{nicefrac}       %
\usepackage{microtype}      %
\usepackage{xcolor}         %

\usepackage[pdftex]{graphicx}
\usepackage{multirow}
\usepackage{microtype}
\usepackage{hyperref}
\usepackage{url}
\usepackage{booktabs}
\usepackage{float}
\usepackage[most]{tcolorbox}
\usepackage{tabularx}
\usepackage{tcolorbox}

\usepackage{booktabs}       %
\usepackage{makecell}
\usepackage{multirow}
\usepackage{tabu}
\usepackage{threeparttable}
\usepackage{tabularx}
\usepackage{tabulary}

\usepackage{tikz}
\usetikzlibrary{bayesnet, shapes, arrows, positioning, arrows.meta}
\usepackage{pgfplots}
\usepackage{pgfplotstable}
\pgfplotsset{compat=newest}

\usepackage[normalem]{ulem}
\usepackage{wrapfig}
\usepackage[compatibility=false]{caption}
\usepackage[labelformat=simple]{subcaption}

\usepackage{microtype}      %
\usepackage[most]{tcolorbox}
\usepackage{soul}
\usepackage{paralist}
\usepackage{pifont}
\usepackage{colortbl}

\usepackage{tabularx}
\usepackage{tcolorbox}
\tcbuselibrary{skins, xparse}

\definecolor{lightgray}{gray}{0.9}
\definecolor{subtleblue}{RGB}{30,144,255} %

\newtcolorbox{analysisbox}[1][]{
    enhanced jigsaw,
    colback=white,
    colframe=blue!75!black,
    fonttitle=\bfseries,
    boxsep=5pt,
    left=5pt,
    right=5pt,
    top=5pt,
    bottom=5pt,
    title=#1,
}

\usepackage{hyperref}

\usepackage{url}            %
\usepackage{booktabs}       %
\usepackage{amsfonts}       %
\usepackage{nicefrac}       %
\usepackage{microtype}      %
\usepackage{xcolor}         %

\usepackage{colortbl}
\usepackage{tablefootnote}
\usepackage{graphicx}
\usepackage{multirow}
\usepackage{enumitem}
\usepackage{wrapfig}
\usepackage{subcaption}

\DeclareUnicodeCharacter{00D7}{-}
\DeclareUnicodeCharacter{2212}{-}
\DeclareUnicodeCharacter{00B0}{-}

\newcommand{\name}{\textsc{Natural Plan}}

\title{\name: Benchmarking LLMs on Natural Language Planning}

\author{Huaixiu Steven Zheng \quad Swaroop Mishra \quad Hugh Zhang\thanks{Currently at Scale AI, work done while interning at Google Deepmind}\\ 
\textbf{Xinyun Chen \quad Minmin Chen \quad Azade Nova \quad Le Hou}\\
\textbf{Heng-Tze Cheng \quad Quoc V. Le \quad Ed H. Chi  \quad Denny Zhou}
\\
[1ex]
 Google DeepMind \\
}

\begin{document}

\maketitle

\begin{abstract}
We introduce \name{}, a realistic planning benchmark in natural language containing 3 key tasks: \emph{Trip Planning}, \emph{Meeting Planning}, and \emph{Calendar Scheduling}. We focus our evaluation on the planning capabilities of LLMs with full information on the task, by providing outputs from tools such as Google Flights, Google Maps, and Google Calendar as contexts to the models. This eliminates the need for a tool-use environment for evaluating LLMs on Planning. We observe that \name{} is a challenging benchmark for state of the art models. For example, in \textit{Trip Planning}, GPT-4 and Gemini 1.5 Pro could only achieve 31.1\% and 34.8\% solve rate respectively. We find that model performance drops drastically as the complexity of the problem increases: all models perform below 5\% when there are 10 cities, highlighting a significant gap in planning in natural language for SoTA LLMs. We also conduct extensive ablation studies on \name{} to further shed light on the (in)effectiveness of approaches such as self-correction, few-shot generalization, and in-context planning with long-contexts on improving LLM planning. 

\end{abstract}

\section{Introduction}
Large language models (LLMs) such as Gemini~\citep{team2023gemini} and GPT4~\citep{GPT4} have demonstrated great potentials in solving complex tasks, such as understanding natural language~\citep{winograd1972understanding}, reasoning~\citep{cobbe2021training} and code generation~\citep{chen2021evaluating}. The planning capability of these models however are still far behind human performance~\citep{xie2024travelplanner}. Even though many recent works target to tackle this problem, without realistic benchmarks that reflect real-world applications, it is difficult to make meaningful progress in enhancing these models' planning abilities. 

Planning~\citep{russell2016artificial} has been exhaustively studied in robotics and embodied environments~\citep{LaValle_2006, camacho2007constrained, jiang2019task}.  Automated planning algorithms
have been widely used ~\citep{zhang2015mobile, lagriffoul2018platform, jiang2019multi, ding2020task} to enable planning in such systems. In these settings, planning involves coming up with a sequence of actions which when executed would take an agent from a certain initial state to a desired world state. These systems frequently use planning domain description language (PDDL)~\citep{McDermott_1998_PDDL} or answer set programming (ASP)~\citep{brewka2011answer} as the underlying action language for the planners. Even though recent works have explored the potential of LLM-powered agents in classical planning settings~\citep{silver2023generalized, liu2023llm+, hao2023reasoning, valmeekam2024planning, wang2023promptagent, guan2024leveraging}, planning scenarios are not in practical setting and they are not in natural language as they often contains templates.   

\begin{figure}[H]
\begin{center}
\includegraphics[width=\linewidth]{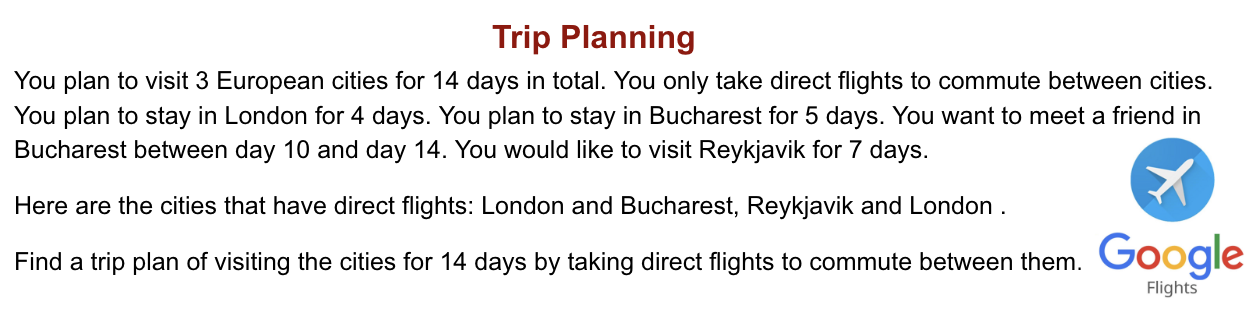}
\caption{Illustrative example of the \emph{Trip Planning} task.}
\label{fig:teaser1}
\end{center}
\end{figure}
\vspace{-5mm}
\begin{figure}[H]
\begin{center}
\includegraphics[width=\linewidth]{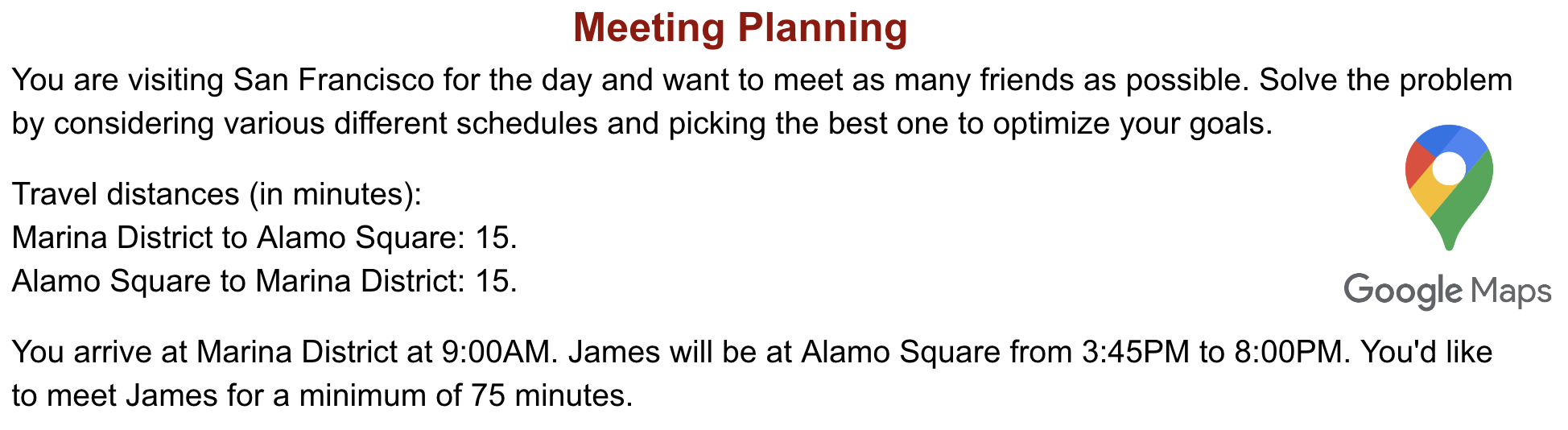}
\caption{Illustrative example of the \emph{Meeting Planning} task.}
\label{fig:teaser2}
\end{center}
\end{figure}
\vspace{-2mm}
\begin{figure}[H]
\vspace{-3mm}
\begin{center}
\includegraphics[width=\linewidth]{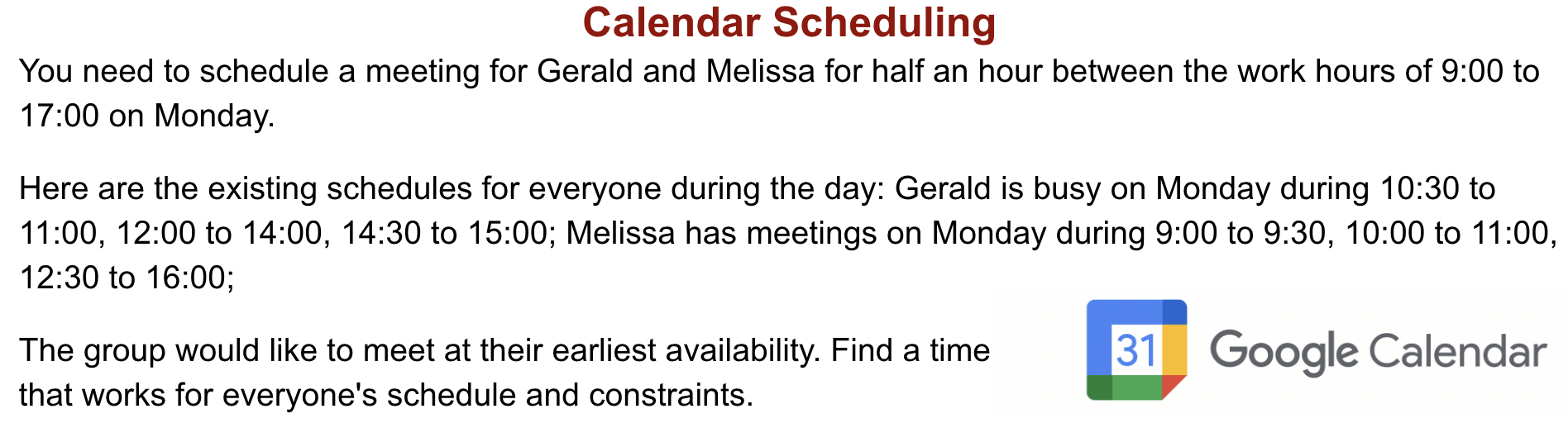}
\caption{Illustrative example of the \emph{Calendar Scheduling} task.}
\label{fig:teaser3}
\end{center}
\end{figure}

In this paper we introduce \name{}, a new benchmark specifically designed to assess how well LLMs can handle planning tasks described in natural language. \name{} includes three different planning tasks i.e., \emph{Trip Planning}, \emph{Meeting Planning}, \emph{Calendar Scheduling}. For each task we gather real information from the existing tools. For example, in \textit{Trip Planning} we query Google Flights API and provide the flight connectivity information between cities in the context. Figures~\ref{fig:teaser1}-~\ref{fig:teaser3} illustrate the tasks along with associated google tools that provide information as context.

Our evaluation shows that \name{} is a hard benchmark for state of the art models. For example, in the \emph{Trip Planning task}, GPT-4 and Gemini 1.5 Pro achieves 31.1\% and 34.8\% performance, while GPT-3.5 is only at 7.3\% and GPT-4o is at 3.7\%. We find that model performance reduces significantly as the complexity of the task increases e.g. an increase in number of people, cities or days in the question. Self-correction does not help models in improving performance. We also observe interesting insights in our easy to hard and hard to easy generalization experiments. Lastly, we conduct study of in-context planning with long context, and show that Gemini 1.5 Pro significantly outperforms other models with $\sim40\%$ accuracy on Trip Planning and $\sim50\%$ on Calendar Scheduling.

\section{\name}
\name{} is a benchmark designed to evaluate capability of LLMs in performing planning tasks expressed in natural language. In this section, we discuss about various categories we have in \name{}, our dataset construction process, quality measures and some statistics.

\subsection{Dataset Categories}
\name{} has 3 planning categories: \emph{Trip Planning}, \emph{Meeting Planning} and \emph{Calendar Scheduling}. Each of these is intended to capture a real world planning task involving some tools such as Google flights. We decouple tool-use from the reasoning task and provide tool outputs in-context to keep focus of \name{} on planning. We believe this will significantly simplify the process of evaluating planning for LLMs.

\subsubsection{\emph{Trip Planning}}

\emph{Trip Planning} is a task focusing on planning a trip itinerary under given constraints. An example prompt and solution is shown in Appendix~\ref{appendix:stats_examples}. The task is to find the itinerary regarding the order of visiting $N$ cities. We add enough constraints such that there is only one solution to the task, which makes the evaluation of the predictions straightforward.

\subsubsection{\emph{Meeting Planning}}

This portion of the benchmark focuses on scheduling meetings under various given constraints. Given $N$ friends, this task provides the meeting times as well as their locations. Appendix~\ref{appendix:stats_examples} demonstrates an example question. The objective is to meet as many friends as possible given the constraints, which include travel time between locations. 

\subsubsection{\emph{Calendar Scheduling}}
\emph{Calendar Scheduling} is a task focusing on scheduling work meetings between multiple people given existing schedules and various constraints. We vary the number of people involved in the meeting and the number of workdays, in order to vary the task difficulties. Appendix~\ref{appendix:stats_examples} demonstrates an example question. 

\subsection{Dataset Construction}
We create \name{} synthetically by using corresponding tool data and creating various constraints. We describe below the specific creation process for each category of task.

\subsubsection{\textit{Trip Planning}}

We create the dataset in the following way: To create a question, we randomly sample $N \in [3, 10]$ cities from the top-48 visited cities in Europe. We also randomly sample the total duration and the number of days spent in each city with $D \in [2, 7]$. We add constraints to the itinerary requiring visiting some cities during particular days, e.g. \textit{meeting a friend in Paris from day 3 to day 5 on the trip}. As a global constraint, we add the preference of commuting between cities using direct flights only, excluding the possibility of visiting city Y after X if there is no direct flight from X to Y. To assist the LLM with the connectivity constraint, we query Google Flights API and provide the flight connectivity information between cities in the context. 

\subsubsection{\textit{Meeting Planning}}
For this dataset category, we randomly sample the person names to meet with and corresponding locations. We also sample time frames showing availability of persons. We add constraints in terms of availability of the subject e.g. you are visiting San Francisco for a day. The goal in this set up is to meet as many friends as possible. In order to make the problem realistic, the time constraints / locations follow real world locations. Travel distances are estimated by querying driving times via the Google Maps API.

\subsubsection{\textit{Calendar Scheduling}}
This portion of the dataset includes tasks to schedule a meeting of either 30 minutes or an hour among various number of attendees in one or multiple of the work days. 
We instantiated two subsets: the first set fixes the meeting on a specific day, e.g.,  Monday while varying the number of attendees in $[2, 7]$; the second set fixes the number of attendees to 2 while varying the number of work days the meeting can be scheduled over in $[1, 5]$. %
Each person involved in the meeting comes with their existing meeting schedule during the day (first set) or the week (second set), which was created by adding random meetings of 30 minutes to the days until $n$ hours of meeting time is filled. We randomly assign $n$ to each attendee to make sure half of the attendees have busy schedule, meaning more than half of their working hours are spent in meetings, while the other half have light schedule with less than half of the working hours spent in meetings. Consecutive meetings of each attendee are combined into a single block as shown in Figure~\ref{fig:teaser3}. %
Once we have the existing schedules of all attendees, we add two types of constraints: 1) earliest availability; 2) avoid meeting during certain time frame to make sure there is a single solution to the problem. 

\subsection{Statistics}

Table ~\ref{table:task_stats} shows statistics of various tasks across \emph{Trip Planning}, \emph{Meeting Planning}, and \emph{Calendar Scheduling} in \name{}.

\begin{table}
\caption{Statistics of \name{} tasks.}
\label{table:task_stats}
\begin{center}
\begin{tabular}{c|c}
\toprule
Task & \#Test Examples \\
\midrule
\textit{Trip Planning} & 1600 \\
\midrule
\textit{Meeting Planning} & 1000 \\
\midrule
\textit{Calendar Scheduling} & 1000	\\
\bottomrule
\end{tabular}
\end{center}
\end{table}

\section{Experiment}
We conduct various experiments to evaluate state-of-the-art LLMs on the planning tasks in \name{}.

\subsection{Models}
We experiment several state-of-the-art LLMs: GPT-4 (gpt-4-turbo-2024-04-09) \citep{GPT4}, GPT-3.5 (gpt-3.5-turbo-0125), GPT-4o (gpt-4o-2024-05-13), and Gemini 1.5 Flash and Pro \citep{geminiteam2024gemini}.

\subsection{Setups}
\textbf{Few-Shot}\newline
For each task in \name{}, we present 5 examples within the same task as few-shot exemplars to enable the LLM in-context learning~\citep{brown2020language} from the examples (see Appendix~\ref{appendix:prompts} for the few-shot prompt). This setup also ensures that outputs can be easily parsed for scoring.

\textbf{Constraint Complexity Understanding}\newline
To gain a better understanding of the planning ability of LLMs, we vary the difficulty level of each task through a controlled variable. 
We conduct analysis by varying constraints such as number of people, number of cities, number of days etc. to better understand behavior of models across the variation in the complexity of the planning task.

\textbf{Few-Shot Generalization}\newline
We conduct easy-to-hard and hard-to-easy generalization study by providing easier and harder examples as in-context exemplars, respectively. For instance, for easy-to-hard (hard-to-easy) generalization in \textit{Trip Planning}, we provide 5 easier (harder) examples with 2 fewer (more) cities than the ones in the evaluation task instances.

\textbf{Self-Correction}\newline
We evaluate if models can self-correct their mistakes across various categories of tasks in \name{}. Here we prompt the model to identify if there are any mistakes and self correct if necessary.

\textbf{In-Context Planning with Long Context}\newline
We evaluate how well models can leverage the long-context capabilities for in-context learning in planning. We split the test-set into 80\% and 20\%, where 80\% is used as in-context examples while we evaluate the model performance on 20\% of the test-set. We increase the number of few-shot examples used up to 800.

\textbf{Evaluation}\newline
We parse the output from the LLM, and compare it with the golden plan. We compute an exact match (EM) score of the LLM-generated plan being identical to the golden plan. See Appendix~\ref{app:parsing_scoring} for the details of parsing and scoring.

\section{Results and Analysis}

\begin{table}[H]%
\caption{Accuracy of 5 models on \name{}.
}
\label{table:main_results}
\begin{center}
\begin{tabular}{r|c|c|c|c|c}
\toprule
Task & GPT-3.5 & GPT-4 & GPT-4o & Gemini 1.5 Flash & Gemini 1.5 Pro \\
\midrule
\emph{Trip Planning} & 7.3\% & 31.1\% & 3.7\% & 25.6\% & \textbf{34.8\%} \\
\emph{Meeting Planning} & 19.1\% & \textbf{47.0\%} & 45.2\% & 23.9\% & 39.1\%\\
\emph{Calendar Scheduling}  & 19.9\% & 41.2\% & 43.7\% & 34.3\% & \textbf{48.9\%} \\
\bottomrule
\end{tabular}
\end{center}
\end{table}

\subsection{Results}

Table~\ref{table:main_results} shows the performance of various LLMs across tasks in \name{} in the 5-shot setup. We get the following insights from the results.

\textbf{All models perform poorly on \name{}.} Gemini 1.5 Pro performs the best on \textit{Trip Planning} and \textit{Calendar Scheduling} while GPT-4 does the best on \textit{Meeting Planning}. 

In \emph{Trip Planning}, Gemini 1.5 Pro achieves 34.8\% accuracy while GPT-4 is 31.1\%. In particular, we observe that GPT-4o is only at 3.7\%. Further analysis shows that GPT-4o is struggling to understand and respect the flight connectivity and travel date constraints, see Appendix~\ref{app:gpt_4o_trip_planning} for a more detailed analysis.
GPT-4 and GPT-4o achieve 47\% and 45.2\% accuracy on \textit{Meeting Planning} while Gemini 1.5 Pro is 39.1\%. For \textit{Calendar Scheduling}, Gemini 1.5 Pro outperforms GPT-4 and GPT-4o at 48.9\%.
 
Comparing the three tasks in \name{}, \emph{Trip Planning} is the hardest among them with the lowest best accuracy of 34.8\%. Note all models have accuracy below 50\% on all three tasks in \name{}, highlighting that planning in natural language is still a very challenging task even for the state of the art models, despite that all tool-use information needed is provided in the context.

\subsection{Constraint Complexity}
\label{sec:constraint_complexity}

\begin{figure}[H]
\begin{center}
\includegraphics[width=1.0\linewidth]{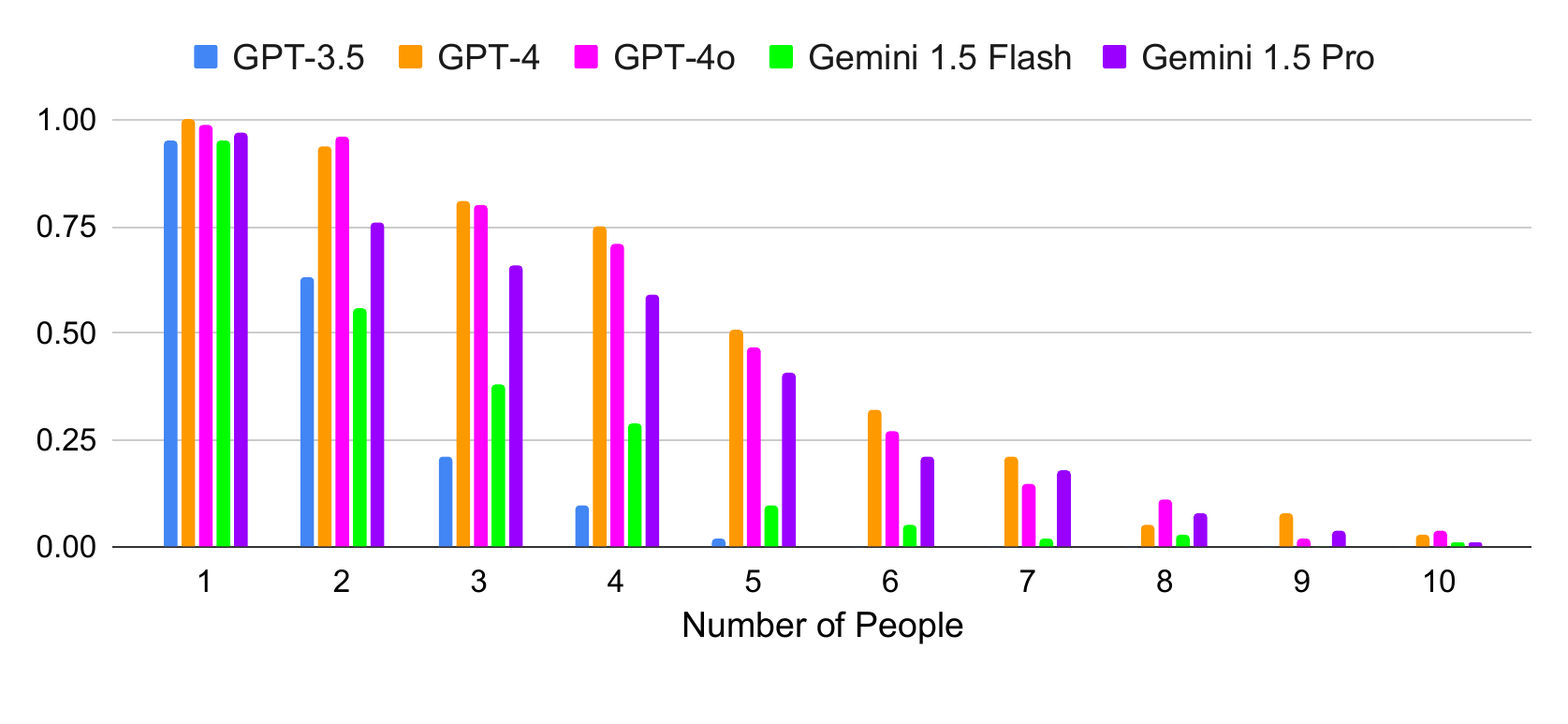}
\vspace{-5mm}
\caption{Constraint complexity analysis of the \textit{Meeting Planning} task in \name{} as a function of number of people.}
\label{fig:meeting_planning_em_vs_people}
\end{center}
\end{figure}

\begin{figure}[H]
\begin{center}
\includegraphics[width=1.0\linewidth]{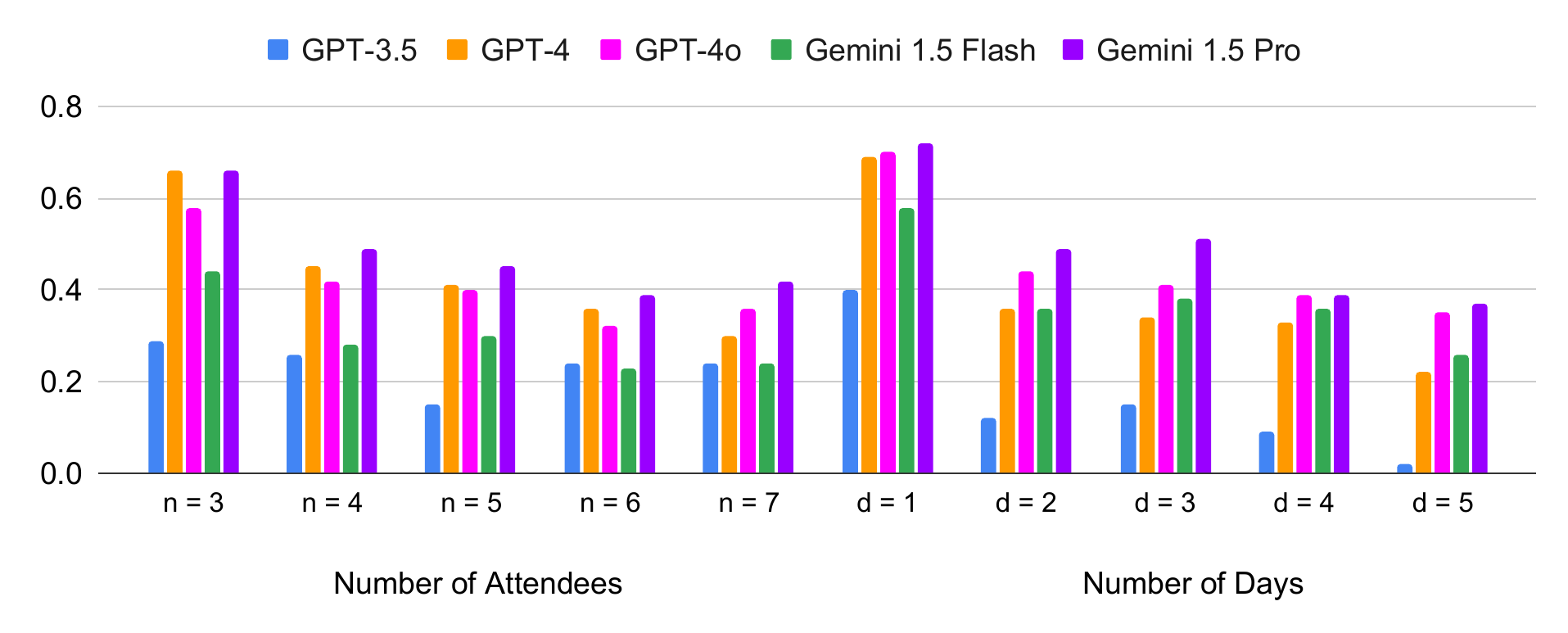}
\vspace{-5mm}
\caption{Constraint complexity analysis of the \textit{Calendar Scheduling} task as a function of the number of people involved to meet during a single day (left), and number of working days to choose a meeting time from between two attendees (right).}
\label{fig:calendar_scheduling_em_vs_people}
\end{center}
\end{figure}

Figures~\ref{fig:trip_plan_em_vs_cities}-~\ref{fig:calendar_scheduling_em_vs_people} show the performance variation of models across variation in number of cities in \textit{Trip Planning}, number of people in \textit{Meeting Scheduling}, and number of attendees/days in \textit{Calendar Scheduling} . We observe the following key insights:
\begin{figure}[H]
\begin{center}
\includegraphics[width=1.0\linewidth]{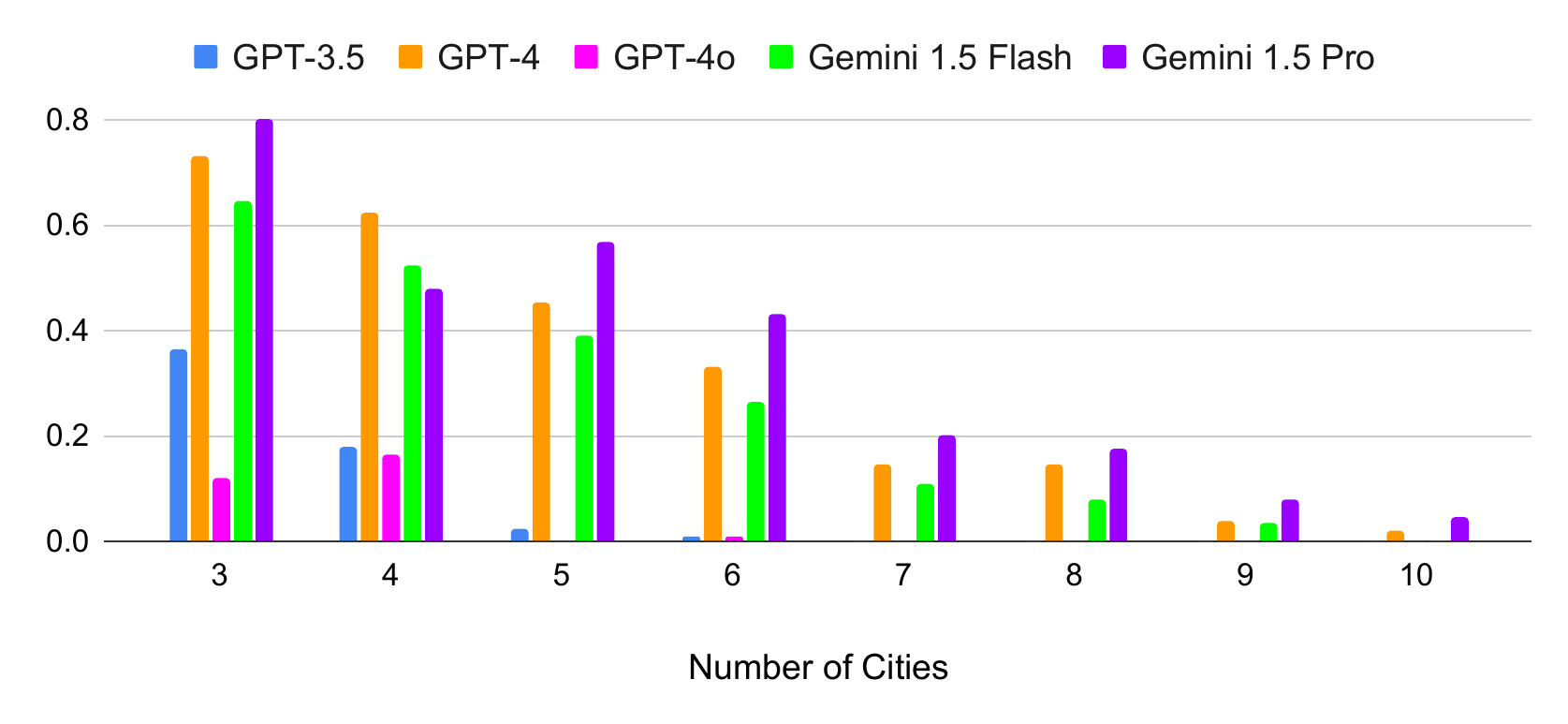}
\vspace{-5mm}
\caption{Constraint complexity analysis of the \textit{Trip Planning} task in \name{} as a function of number of cities.}
\label{fig:trip_plan_em_vs_cities}
\end{center}
\end{figure}
\vspace{-10mm}

\textbf{Model performance reduces significantly as the task complexity increases as number of cities/number of people/number of attendees/number of days increase.} For \textit{Trip Planning}, Figure~\ref{fig:trip_plan_em_vs_cities} shows that models start to fail with more cities involved in the planning. In \textit{Trip Planning}, all five models perform below 5\% when there are 10 cities. For \textit{Meeting Planning}, we evaluate scenarios with the number of people to meet from 1 to 10. Figure~\ref{fig:meeting_planning_em_vs_people} again shows that the performance drops sharply with more people to meet. Beyond 8 people, all models have a score lower than 10\%. For \textit{Calendar Scheduling}, the drop of accuracy is not as significant when increasing number of attendees or number of days compared to the other two tasks.

\subsection{Generalization Analysis}
\begin{figure}[H]
\begin{center}
\includegraphics[width=1.0\linewidth]{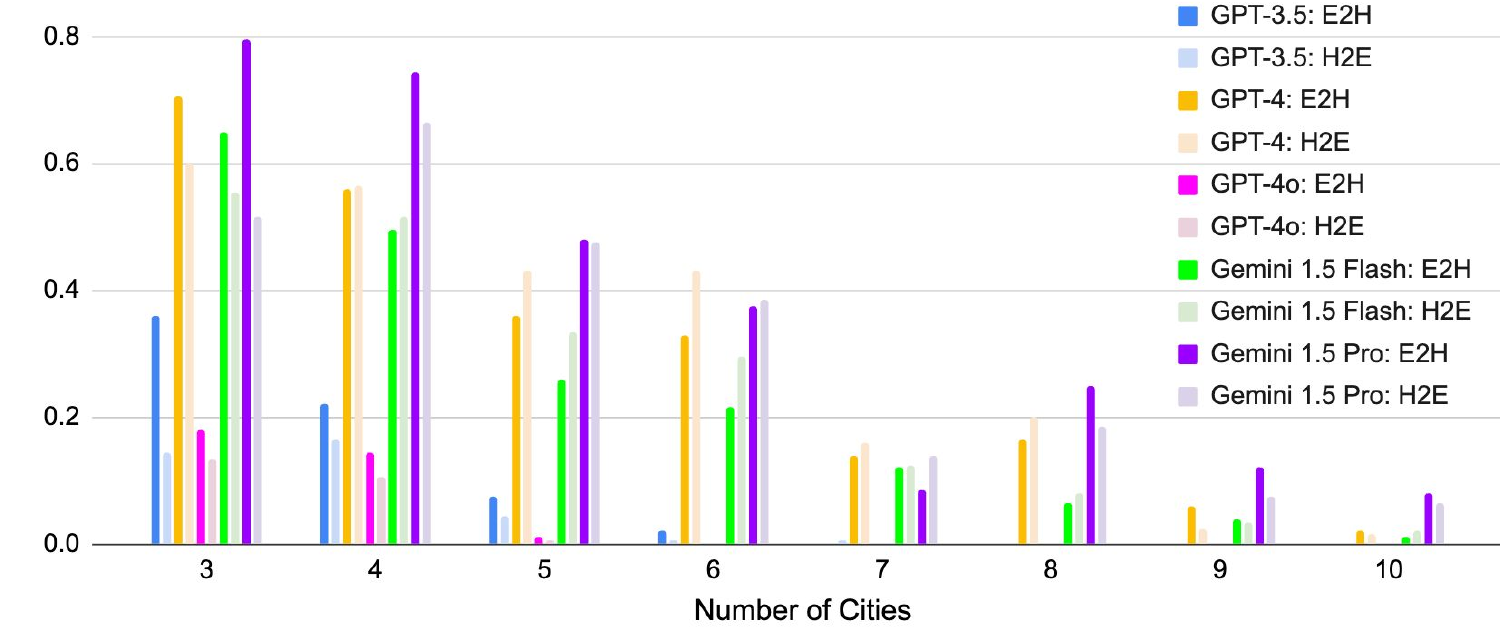}
\caption{Model performance on the \textit{Trip Planning} task as a function of number of cities for few-shot generalization ablations: E2H=easy-to-hard, H2E=hard-to-easy.}
\label{fig:trip_plan_fewshot_gen}
\end{center}
\end{figure}

Figure~\ref{fig:trip_plan_fewshot_gen} shows the summary of our easy-to-hard and hard-to-easy generalization experiments. For easy-to-hard (hard-to-easy) generalization, we present 5 easier (harder) examples with 2 fewer (more) cities than the task, to test whether LLMs can generalize from easier (harder) in-context learning examples. We choose \textit{Trip Planning} to conduct our ablations on given that it is the most difficult task among \name{} tasks.

\textbf{In general, we observe that easy-to-hard performs better than hard-to-easy generalization across model families.} Section~\ref{sec:constraint_complexity} shows models perform worse when the task complexity is higher. This result indicates that understanding and making use of hard demonstrations in in-context learning is still challenging even for state of the art LLMs.  However, we do observe that for GPT-4 and Gemini 1.5 Flash, the trend is reversed as the task complexity increases: for $N=5\sim8$, hard-to-easy demonstrations work better than easy-to-hard.

\subsection{Self-Correction}

\begin{figure}[H]
\begin{center}
\includegraphics[width=1.0\linewidth]{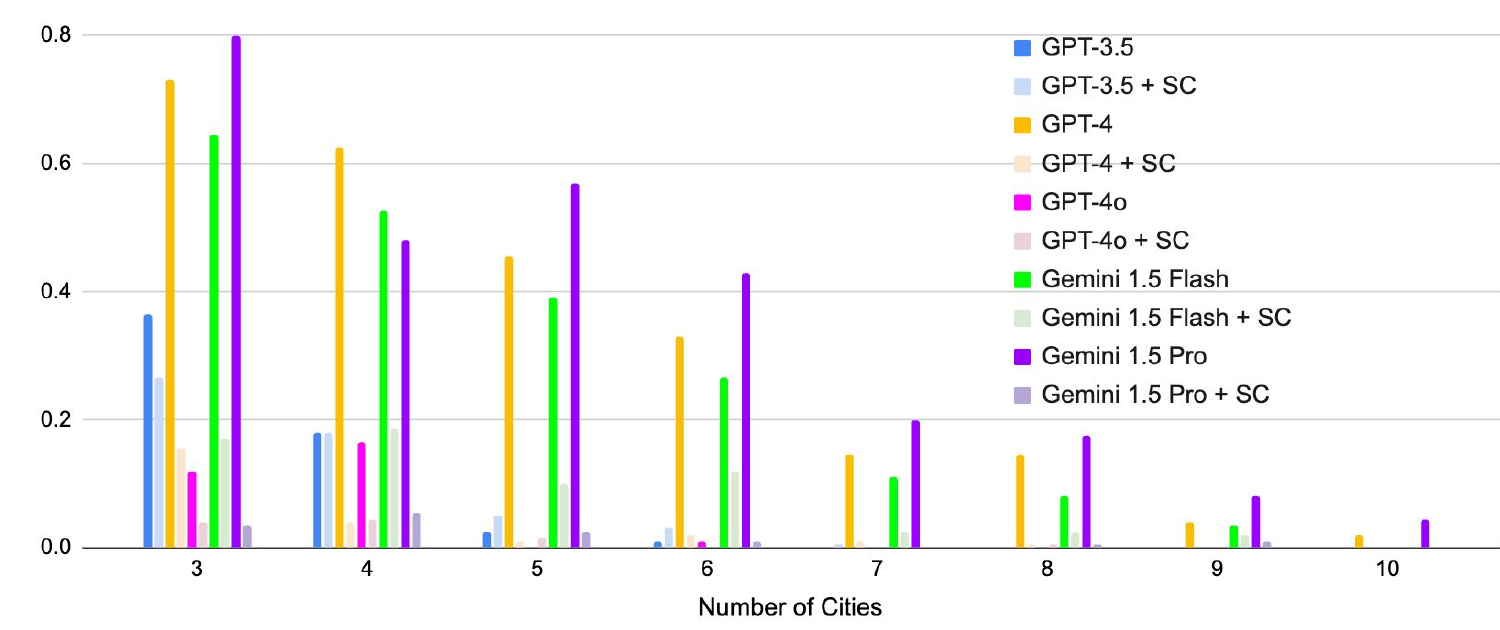}
\caption{Accuracy of \textit{Trip Planning} as a function of number of cities for self-correction ablations. SC=self-correction}
\label{fig:trip_plan_sc}
\end{center}
\end{figure}
Figure~\ref{fig:trip_plan_sc} shows the results of performing self-correction on model predictions in the \textit{Trip Planning} task. We prompt the LLM to check whether the generated plan is correct or not, and if determined incorrect, fix the plan (see Appendix~\ref{appendix:prompts} for the prompt). 

\textbf{Self-correction leads to significant model performance drop across all models. Interestingly, the stronger models such as GPT-4 and Gemini 1.5 Pro suffer bigger loss compared to GPT-3.5.} This aligns well with previous findings that self-correction could lead to performance degradation on Reasoning \citep{huang2023large}. The stronger instruction-tuned models like GPT-4 and Gemini 1.5 Pro could be over-confident towards correcting its own solutions when prompted, see Appendix~\ref{appendix:sc_example} for examples of self-correction.

\subsection{In-Context Planning with Long Context}
\begin{figure}[H]
\begin{center}
\includegraphics[width=1.0\linewidth]{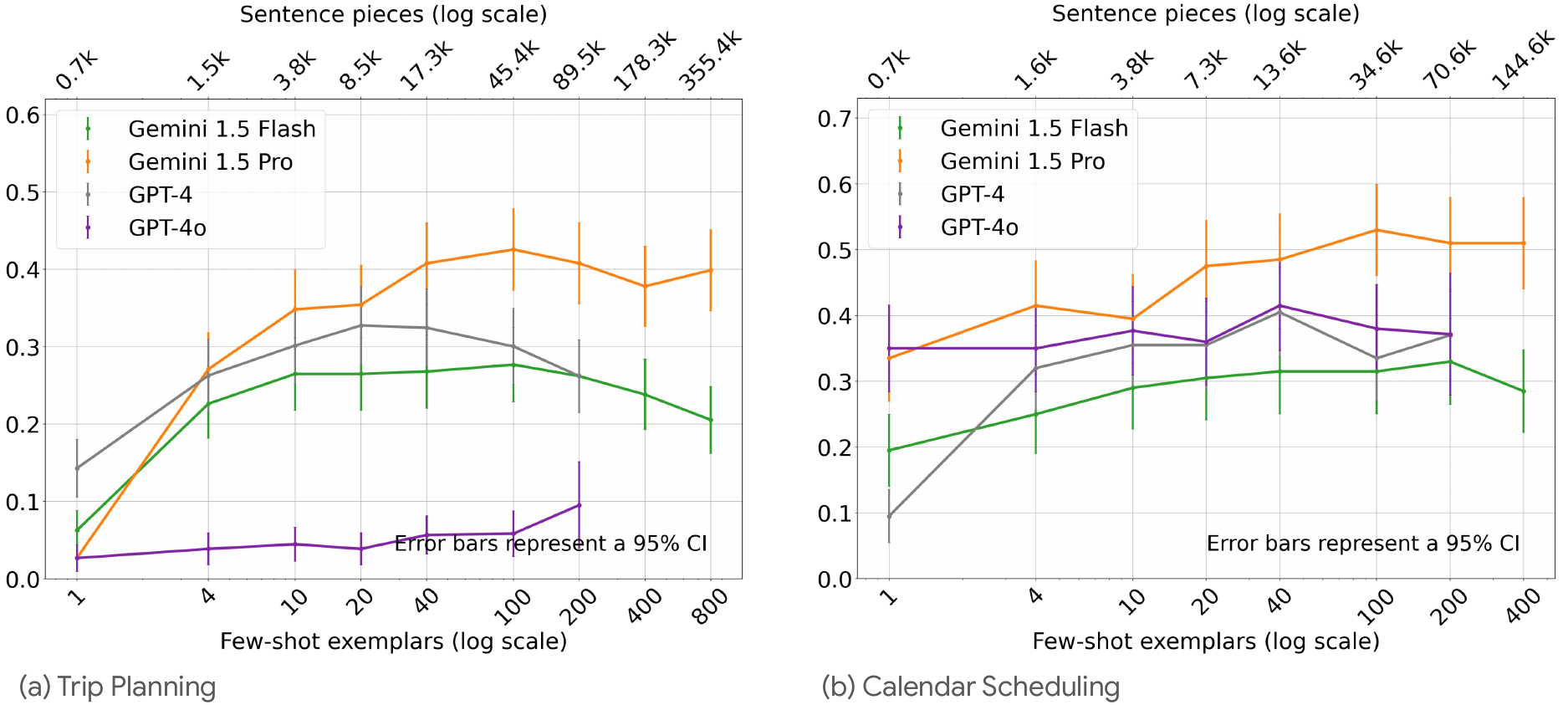}
\caption{In-Context planning with long context for \textit{Trip Planning} and \textit{Calendar Scheduling} tasks. Gemini 1.5 Pro shows strong long-context learning on Planning compared to other models.}
\label{fig:in_context_planning}
\end{center}
\end{figure}

In this section, we study the possibility of in-context planning by leveraging the long context capabilities of recent models \citep{geminiteam2024gemini}. Figure~\ref{fig:in_context_planning} shows the performance of four models up to 800 shots.

\textbf{Due to the strong long context capabilities, Gemini Pro 1.5 is able to leverage more in-context examples up to 355K tokens, still showing steady improvements.} For instance, in \textit{Trip Planning}, increasing number of shots from 1 to 800 brings up the accuracy of Gemini Pro 1.5 from 2.7\% to 39.9\%. In the mean time, both GPT-4 and Gemini 1.5 Flash peak at 20-shot and start to show performance degradation. The same is observed in \textit{Calendar Scheduling} where Gemini 1.5 Pro continually improves up to 400 shots. These results show the promise of in-context planning where the long-context capabilities enable LLMs to leverage further context to improve Planning.

\section{Related Works}
Planning empowers intelligent agents to look ahead and proactively determine a course of action to reach objectives~\citep{mccarthy1963situations, russell2016artificial}. Automated/classical planning is a subarea of AI that allows sequential decision-making systems to choose a sequence of actions using which they complete a (set of) task(s) autonomously~\citep{GhallabNauTraverso04}. Most of the planning formalisms use a Planning Domain Definition Language (PDDL)~\citep{McDermott_1998_PDDL}, where it separates the environment dynamics (domain definition) from the environment configuration (problem definition).  Domain-independent planners like Fast-Forward~\citep{hoffmann2001}, Fast-Downward~\citep{Helmert06thefast}, etc. take these domain and problem definitions as input and generate a sequence of actions that takes a planning agent from a given initial state to a state satisfying a goal condition (also given as input with the problem definition). Such classical planning methods
have been widely adopted in robots and embodied environments~\citep{camacho2007constrained, jiang2019task}. 
Most planning benchmarks~\footnote{https://www.icaps-conference.org/competitions/} use similar formulations with variations like probabilities~\citep{younes2005first}, numeric fluents~\footnote{https://ipc2023-numeric.github.io/}, temporal actions, etc. as benchmarks. The main issue with such benchmarks is that they are not expressed in natural language and hence need experts to handcraft the domains and problems.

Recently prompting LLMs to extract common sense knowledge gained attention~\citep{huang2022inner, singh2023progprompt, ding2023task}. Effectiveness of LLMs in generating plans has been studied by~\cite{valmeekam2024planning, hao2023reasoning, guan2024leveraging}. While~\cite{valmeekam2024planning, guan2024leveraging} translate natural language instructions into the executable programming languages, such as PDDL, and runs classical planning algorithms,~\cite{hao2023reasoning} uses the LLM as both a world model and a reasoning agent, enhanced by a Monte Carlo Tree Search algorithm for efficient exploration.

Even though these works target to improve the planning capability of LLMs, these are not able to perform well in more complex and realistic settings, closer to those in which humans operate~\citep{valmeekam2022large}. TravelPlanner~\citep{xie2024travelplanner} provides a sandbox with various tools to evaluate language model capabilities in performing planning via agents. Our work \name{} provides a holistic benchmark in natural language to evaluate the planning capability of models in real world tasks.

\section{Conclusion}

We introduce \name{} to evaluate the capability of LLMs to plan on real world tasks given in natural language. \name{} covers three domains: \textit{Trip Planning}, \textit{Calendar Scheduling}, and \textit{Meeting Planning}. The tasks also contain API results from various tools such as Google Calendar, Google Flights and Google Maps mimicking real world planning tasks with tools. We find that \name{} is very hard for state of the art models to solve. Models increasingly struggle as the complexity of the problem increases, e.g. an increase in number of people, cities or days in the question. However, our in-context planning experiments demonstrate the potential promise of leveraging the long-context capabilities to boost LLMs' planning ability.

\subsubsection*{Acknowledgments}
We acknowledge careful review and feedback from Uri Alon and Aleksandra Faust of Google DeepMind on the draft of this paper.

\bibliography{main}
\bibliographystyle{plainnat}

\appendix

\section{\name{} Examples}
\label{appendix:stats_examples}

Below is one example pair of the prompt and solution for each task in \name{}. 

\begin{analysisbox}[\textit{Trip Planning}]

\textbf{Prompt:}\\ You plan to visit 3 European cities for 14 days in total. You only take direct flights to commute between cities. You would like to visit Florence for 6 days. You want to meet a friend in Florence between day 9 and day 14. You would like to visit Barcelona for 5 days. You would like to visit Helsinki for 5 days. \\
    
    Here are the cities that have direct flights: Barcelona and Florence, Helsinki and Barcelona. \\
    
    Find a trip plan of visiting the cities for 14 days by taking direct flights to commute between them. \\

\textbf{Solution:} \\ Here is the trip plan for visiting the 3 European cities for 14 days:
    
    **Day 1-5:** Arriving in Helsinki and visit Helsinki for 5 days.\\
    **Day 5:** Fly from Helsinki to Barcelona. \\
    **Day 5-9:** Visit Barcelona for 5 days. \\
    **Day 9:** Fly from Barcelona to Florence. \\
    **Day 9-14:** Visit Florence for 6 days.

\end{analysisbox}

\begin{analysisbox}[\textit{Meeting Planning}]

\textbf{Prompt:} \\
You arrive at SOMA (South of Market) at 9:00AM. Joseph will be at Alamo Square from 4:15PM to 6:30PM. You'd like to meet Joseph for a minimum of 105 minutes. Andrew will be at Haight-Ashbury from 4:30PM to 6:15PM. You'd like to meet Andrew for a minimum of 30 minutes. John will be at Nob Hill from 7:15AM to 11:00AM. You'd like to meet John for a minimum of 75 minutes.\\

It takes 27 minutes to travel between Sunset District and Nob Hill via car. It takes 18 minutes to travel between Sunset District and Alamo Square via car. It takes 11 minutes to travel between Sunset District and Golden Gate Park via car. It takes 25 minutes to travel between Nob Hill and Sunset District via car. It takes 11 minutes to travel between Nob Hill and Alamo Square via car. It takes 18 minutes to travel between Nob Hill and Golden Gate Park via car. It takes 16 minutes to travel between Alamo Square and Sunset District via car. It takes 11 minutes to travel between Alamo Square and Nob Hill via car. It takes 9 minutes to travel between Alamo Square and Golden Gate Park via car. It takes 10 minutes to travel between Golden Gate Park and Sunset District via car. It takes 19 minutes to travel between Golden Gate Park and Nob Hill via car. It takes 10 minutes to travel between Golden Gate Park and Alamo Square via car.\\

\textbf{Solution:} \\
You start at SOMA (South of Market) at 9:00AM. You travel to Nob Hill in 10 minutes and arrive at 9:10AM. You meet John for 75 minutes from 9:10AM to 10:25AM. You travel to Alamo Square in 11 minutes and arrive at 10:36AM. You wait until 4:15PM. You meet Joseph for 105 minutes from 4:15PM to 6:00PM. 

\end{analysisbox}

\begin{analysisbox}[\textit{Calendar Scheduling}]

\textbf{Prompt:} \\
You need to schedule a meeting for Harold and Patrick for half an hour between the work hours of 9:00 to 17:00 on Monday. \\

Harold's calendar is wide open the entire day.
Patrick is busy on Monday during 9:00 to 9:30, 10:30 to 12:00, 12:30 to 13:30, 14:00 to 14:30, 15:00 to 16:30; \\

Find a time that works for everyone's schedule and constraints. \\

\textbf{Solution:} \\
Here is the proposed time: Monday, 9:30 - 10:00 

\end{analysisbox}

\section{Prompts}
\label{appendix:prompts}
Here is the 5-shot prompt for inference of \textit{Trip Planning}. Similar prompts are used for \textit{Meeting Planning}, and \textit{Calendar Scheduling}.
\begin{analysisbox}[Standard 5-Shot Prompt for \textit{Trip Planning}]
\textbf{Prompt:} \\
You are an expert at planning trips. You are given a few constraints regarding the cities to visit and the durations of staying at each city. You are also given the flight information between the cities. Here are a few example tasks and solutions:

TASK: $<$task1$>$ \\
    
SOLUTION: $<$solution1$>$ \\
...\\
TASK: $<$task5$>$ \\
    
SOLUTION: $<$solution5$>$ \\

TASK: $<$task$>$ \\
    
SOLUTION:
\end{analysisbox}

Here is the prompt used for easy-to-hard and hard-to-easy generalization ablations of \textit{Trip Planning}:
\begin{analysisbox}[Easy-to-Hard/Hard-to-Easy Prompt for \textit{Trip Planning}]
\textbf{Prompt:} \\
You are an expert at planning trips. You are given a few constraints regarding the cities to visit and the durations of staying at each city. You are also given the flight information between the cities. Here are a few example tasks and solutions:

TASK: $<$easy/hard task1$>$ \\
    
SOLUTION: $<$solution1$>$ \\
...\\
TASK: $<$easy/hard task5$>$ \\
    
SOLUTION: $<$solution5$>$ \\

TASK: $<$task$>$ \\
    
SOLUTION:
\end{analysisbox}

Here is the self-correction prompt used for \textit{Trip Planning} self-correction ablation:\\

\begin{analysisbox}[Self-Correction Prompt for \textit{Trip Planning}]

\textbf{Prompt:} \\
You are an expert at planning trips. You are given a TASK of Trip Planning request, and a proposed SOLUTION. Your job is to identify any error in the SOLUTION, correct the error and generate a new solution. If there is no error, simple output the proposed SOLUTION.\\

TASK: $<$task$>$ \\
    
SOLUTION: $<$model output$>$ \\

Please self correct the solution if there is any error and output your solution in the following format and keep the NEW SOLUTION in the same format as SOLUTION:\\
**ERROR IDENTIFIED:**\\
**NEW SOLUTION:**
\end{analysisbox}

\section{Plan Parsing and Scoring}
\label{app:parsing_scoring}
In this section, we describe the details on how to parse the plans generated by models, and score the exact-match score against the ground-truth plans. We rely on the few-shot exemplars to make sure the generated plans from models follow the same template solutions provided in few-shot examples. Then, we rely on regex expressions to extract the exact date/place/time information to compare against the ground-truth plans. For example, in \textit{Trip Planning}, we use a regex expression \textit{r".*Day (\textbackslash d+).*from (\textbackslash w+) to (\textbackslash w+)"} to parse the date D of flight from city X to Y. We do similar regex parsing to the other two tasks. For exact match score, only if all the information (e.g. D, X and Y) matches, we score the plan to be correct.

\section{Analysis of GPT-4o on \textit{Trip Planning}}
\label{app:gpt_4o_trip_planning}
We sample 10 examples where GPT-4o made an error on \textit{Trip Planning}, and annotate the type of errors GPT-4o made. All of the errors are due to the fact that GPT-4o failed to understand and respect the constraints in the task. 7/10 are failures in respecting the Flight Connectivity constraints while the other 3/10 are due to Travel Date constraint failures. Below shows an example for both cases.

\begin{analysisbox}[GPT-4o Error on \textit{Trip Planning}: Flight Connectivity]
\textbf{Task:} \\
You plan to visit 3 European cities for 14 days in total. You only take direct flights to commute between cities. You would like to visit Florence for 6 days. You want to meet a friend in Florence between day 9 and day 14. You would like to visit Barcelona for 5 days. You would like to visit Helsinki for 5 days.\\

Here are the cities that have direct flights:
Barcelona and Florence, Helsinki and Barcelona.\\

Find a trip plan of visiting the cities for 14 days by taking direct flights to commute between them.\\

\textbf{Solution:} \\
Here is the trip plan for visiting the 3 European cities for 14 days:\\

**Day 1-5:** Arriving in Helsinki and visit Helsinki for 5 days.\\
**Day 5:** Fly from Helsinki to Barcelona.\\
**Day 5-9:** Visit Barcelona for 5 days.\\
**Day 9:** Fly from Barcelona to Florence.\\
**Day 9-14:** Visit Florence for 6 days.\\
    
\textbf{GPT-4o Output:} \\
"Here is the trip plan for visiting the 3 European cities for 14 days:\\

**Day 1-5:** Arriving in Barcelona and visit Barcelona for 5 days.\\
**Day 5:** Fly from Barcelona to Helsinki.\\
**Day 5-10:** Visit Helsinki for 5 days.\\
\textcolor{red}{**Day 10:** Fly from Helsinki to Florence.}\\
**Day 10-14:** Visit Florence for 6 days.\\

\end{analysisbox}

\begin{analysisbox}[GPT-4o Error on \textit{Trip Planning}: Travel Date]
\textbf{Task:} \\
You plan to visit 3 European cities for 8 days in total. You only take direct flights to commute between cities. You would like to visit Dubrovnik for 3 days. You would like to visit Berlin for 4 days. During day 1 and day 4, you have to attend a conference in Berlin. You would like to visit Munich for 3 days.\\

Here are the cities that have direct flights:
Munich and Dubrovnik, Berlin and Munich.

Find a trip plan of visiting the cities for 8 days by taking direct flights to commute between them.\\

\textbf{Solution:} \\
"Here is the trip plan for visiting the 3 European cities for 8 days:\\

**Day 1-4:** Arriving in Berlin and visit Berlin for 4 days.\\
**Day 4:** Fly from Berlin to Munich.\\
**Day 4-6:** Visit Munich for 3 days.\\
**Day 6:** Fly from Munich to Dubrovnik.\\
**Day 6-8:** Visit Dubrovnik for 3 days."\\
    
\textbf{GPT-4o Output:} \\
"ere is the trip plan for visiting the 3 European cities for 8 days:

**Day 1-4:** Arriving in Berlin and attend the conference while visiting Berlin for 4 days.
**Day 4:** Fly from Berlin to Munich.\\
\textcolor{red}{**Day 4-7:** Visit Munich for 3 days.\\
**Day 7:** Fly from Munich to Dubrovnik.\\
**Day 7-10:** Visit Dubrovnik for 3 days.}

\end{analysisbox}

\section{Self-Correction Example}
\label{appendix:sc_example}
Here we show an example of self-correction on \textit{Trip Planning}, to illustrate the model behavior after self-correction. The errors made by the models are highlighted in red.\\
\begin{analysisbox}[Self-Correction Example]

\textbf{Task:} \\
You plan to visit 3 European cities for 14 days in total. You only take direct flights to commute between cities. You would like to visit Florence for 6 days. You want to meet a friend in Florence between day 9 and day 14. You would like to visit Barcelona for 5 days. You would like to visit Helsinki for 5 days. \\
    
    Here are the cities that have direct flights: Barcelona and Florence, Helsinki and Barcelona. \\
    
    Find a trip plan of visiting the cities for 14 days by taking direct flights to commute between them. \\
\textbf{Solution:} \\
Here is the trip plan for visiting the 3 European cities for 14 days:
    
    **Day 1-5:** Arriving in Helsinki and visit Helsinki for 5 days.\\
    **Day 5:** Fly from Helsinki to Barcelona. \\
    **Day 5-9:** Visit Barcelona for 5 days. \\
    **Day 9:** Fly from Barcelona to Florence. \\
    **Day 9-14:** Visit Florence for 6 days. \\
    
\textbf{GPT-3.5 Output:} \\
**ERROR IDENTIFIED:** The error in the solution is that the friend's visit in Florence is scheduled between day 10 and day 14, instead of between day 9 and day 14 as requested.\\
**NEW SOLUTION:** Here is the corrected trip plan for visiting the 3 European cities for 14 days:\\
**Day 1-5:** Arriving in Helsinki and visit Helsinki for 5 days.\\
**Day 5:** Fly from Helsinki to Barcelona. \\
\textcolor{red}{**Day 5-10:** Visit Barcelona for 5 days.} \\
\textcolor{red}{**Day 10:** Fly from Barcelona to Florence.} \\
\textcolor{red}{**Day 10-15:** Visit Florence for 6 days.} \\
\textbf{GPT-4 Output:} \\
**ERROR IDENTIFIED:** The proposed solution has an error in the total number of days spent. The total number of days spent in the proposed solution is 15 days (5 days in Helsinki, 5 days in Barcelona, and 6 days in Florence), which is more than the 14 days available for the trip. \\
**NEW SOLUTION:** Here is the corrected trip plan for visiting the 3 European cities for 14 days: \\
\textcolor{red}{**Day 1-4:** Arriving in Helsinki and visit Helsinki for 4 days.} \\
**Day 5:** Fly from Helsinki to Barcelona. \\
**Day 5-9:** Visit Barcelona for 5 days. \\
\textcolor{red}{**Day 10:** Fly from Barcelona to Florence.} \\
\textcolor{red}{**Day 10-14:** Visit Florence for 5 days.} \\
\textbf{Gemini Pro 1.5 Output:} \\
"**ERROR IDENTIFIED:** The proposed solution has overlapping days. For example, Day 5 is used for both visiting Helsinki and flying to Barcelona. Additionally, the solution doesn't account for the fact that there are no direct flights between Helsinki and Florence.\\

**NEW SOLUTION:** Here is the trip plan for visiting the 3 European cities for 14 days:\\
\textcolor{red}{
**Day 1-5:** Arriving in Barcelona and visit Barcelona for 5 days.\\
**Day 6:** Fly from Barcelona to Florence.\\
**Day 6-11:** Visit Florence for 6 days (meeting the friend during this time).\\
**Day 12:** Fly from Florence to Barcelona.\\
**Day 12-13:**  Spend an extra day in Barcelona.\\
**Day 14:** Fly from Barcelona to Helsinki. }
\end{analysisbox}

\end{document}